\pdfoutput=1
\documentclass[11pt]{article}

\usepackage{acl}

\usepackage{times}
\usepackage{latexsym}

\usepackage[T1]{fontenc}
\usepackage[utf8]{inputenc}

\usepackage{graphicx}
\usepackage{xspace}
\usepackage{multicol}
\usepackage{enumitem}
\setlist{itemsep=1pt,topsep=1pt}

\newcommand{\dataset}{AustenAlike\xspace}
\newcommand{\fanfic}{FanfictionNLP\xspace}
\newcommand{\booknlp}{BookNLP\xspace}
\newif\ifdraft\draftfalse % set to \draftfalse to hide
\definecolor{wellesleyblue}{RGB}{0, 39, 118}
\definecolor{swarthmoregarnet}{RGB}{99, 25, 25}

\ifdraft{\pagenumbering{arabic}}\fi
\newif\ifsub\subfalse

\usepackage{microtype}

\title{Evaluating Computational Representations of Character:\\ An Austen Character Similarity Benchmark}

\author{Funing Yang \and Carolyn Jane Anderson\\
  Wellesley College \\
  Wellesley, MA \\
  \texttt{carolyn.anderson@wellesley.edu} \\}

\begin{document}
\maketitle
\begin{abstract}

Several systems have been developed to extract information about characters to aid computational analysis of English literature. We propose character similarity grouping as a holistic evaluation task for these pipelines. We present \dataset, a benchmark suite of character similarities in Jane Austen's novels. Our benchmark draws on three notions of character similarity: a structurally defined notion of similarity; a socially defined notion of similarity; and an expert defined set extracted from literary criticism.

We use \dataset to evaluate character features extracted using two pipelines, \booknlp and \fanfic. We build character representations from four kinds of features and compare them to the three \dataset benchmarks and to GPT-4 similarity rankings. We find that though computational representations capture some broad similarities based on shared social and narrative roles, the expert pairings in our third benchmark are challenging for all systems, highlighting the subtler aspects of similarity noted by human readers.
\end{abstract}

\section{Introduction}

There is growing interest in using computational techniques to analyze works of literary fiction. Several systems have been developed to automatically extract information about characters from English literary text~\citep{bamman-etal-2014-bayesian,yoder-etal-2021-fanfictionnlp}. In this paper, we explore character similarity as a holistic evaluation task for literary pipelines. We use character similarity to explore the information about characters that is captured by the different kinds of features these pipelines extract: their events, utterances, and attributes.

Because characters can be similar along multiple axes, we construct a multi-part benchmark, \dataset, that uses three different notions of character similarity to group characters in Jane Austen's novels. The first is a structurally defined notion of similarity to group Austen's characters: characters are similar if they fill similar narrative roles. The second is a socially defined notion of similarity: characters are similar if they share demographic features. The final benchmark takes a wisdom-of-the-crowd approach, but with an expert crowd: we extract comparisons of characters from four decades of \textit{Persuasions}, a journal dedicated to the analysis of Austen's work. Figure \ref{fig:ex} shows an example of how these three views of character similarity can lead to different comparisons.

We use \dataset to explore how much information about characters is captured by the different kinds of features that literary pipelines extract. We extract character events, quotes, modifiers, and assertions using the BookNLP \citep{bamman-etal-2014-bayesian,sims-etal-2019-literary} and FanfictionNLP \citet{yoder-etal-2021-fanfictionnlp} pipelines. We build character representations using contextualized embeddings of these features, and compare how well these representations align with the three sets of character groupings in the \dataset benchmarks. We also compare a non-feature-based approach by extracting similarity judgments from ChatGPT. 

\begin{figure}
\textbf{James Morland from \textit{Northanger Abbey}}\\
\textit{Sibling to heroine and single 20-year-old male clergy with income of \pounds 400/year}

\textbf{Social Pairings:} Charles Hayter, Edward Ferrars, Robert Martin

\textbf{Narrative Role Pairings:} Isabella Knightley, John Dashwood, Margaret Dashwood, Susan Price, William Price, Elizabeth Elliot, Mary Musgrove, Jane Bennet, Mary Bennet, Kitty Bennet, Lydia Bennet

\textbf{Expert Pairings:} Edmund Bertram, Edward Ferrars, Henry Tilney, Philip Elton
    \caption{Example character from \dataset}
    \label{fig:ex}
\end{figure}

Our results show that event- and assertion-based representations capture more information about character similarity than quote-based representations. Overall, however, we show that though computational representations capture some broad social and narratological similarities, there is a wide gap between the similarities they capture and the more nuanced similarities highlighted in our wisdom-of-the-expert-crowd benchmark. The best feature-based representations exhibit only medium correlations with expert rankings of character similarity, and GPT-4 lists the expert-identified most similar character in a top ten similarity list only half of the time. \dataset illustrates how much work remains to achieve nuanced computational representations of literary characters.

\section{Representing Literary Characters}

There is a growing interest in applying computational methods to analyze literary fiction, both in analyses of large collections (\textit{distant reading}~\cite{moretti})~\cite{inproceedings,jayannavar-etal-2015-validating, milli-bamman-2016-beyond} and of individual authors and works~\cite{agarwal-etal-2013-automatic,wang-iyyer-2019-casting,liebl-burghardt-2020-shakespeare}. Though these projects range in scope, they share a foundation of feature extraction: literary evidence must be identified before it can be interpreted. 

To facilitate computational analysis, a number of pipelines for extracting features from literary text have been developed~\citep{bamman-etal-2014-bayesian,sims-etal-2019-literary,yoder-etal-2021-fanfictionnlp,ehrmanntraut_llpro_2023}. In this paper, we focus specifically on features related to literary characters.

\paragraph{Character mentions}

The first step in computational studies of character is to identify character mentions using named entity recognition and coreference resolution. There is a large body of existing work on these tasks~\citep{vala-etal-2015-mr, Brooke2016BootstrappedTN, roesiger-teufel-2014-resolving,jahan-finlayson-2019-character} given their complexity in a literary setting and their importance for downstream tasks. 

Some pipelines further disambiguate character references in a \textit{character clustering} step. BookNLP is a pipeline trained on data from LitBank, which provides annotated training data drawn from 19th- and early 20th-century English fiction, including annotations for named entity recognition~\citep{bamman-etal-2019-annotated} and coreference resolution~\citep{bamman-etal-2020-annotated}. FanfictionNLP  is a similar pipeline that is trained on and tailored to fanfiction.

\paragraph{Character features}

Once character mentions have been identified, the surrounding text can be used to extract information related to characters.

Some previous work focuses on character personality traits and emotions~\cite{kim-klinger-2019-frowning,flekova-gurevych-2015-personality,pizzolli-strapparava-2019-personality}. The pipelines we study target more general descriptions: for \fanfic, \textit{assertions}, descriptions of physical and mental attributes; for \booknlp, modifiers and possessions.

What characters do and say is also of interest. Although quote attribution remains a challenging task with a number of approaches~\citep{he-etal-2013-identification,almeida-etal-2014-joint,muzny-etal-2017-two}, it is useful for analyzing both the content and style of characters' speech~\citep{dinu-uban-2017-finding, vishnubhotla-etal-2019-fictional}. \booknlp extracts both events and quotes, while \fanfic extracts only quotes.

There is also much work on mapping and analyzing relationships between characters~\cite{elson-etal-2010-extracting,lee-yeung-2012-extracting,jayannavar-etal-2015-validating,agarwal-etal-2013-automatic,wohlgenannt-etal-2016-extracting,iyyer-etal-2016-feuding,labatut-extraction-2019}.

\paragraph{Character models}

Once character features are extracted, they can be used to build computational representations of characters. Some work seeks to classify characters into types~\cite{chambers-jurafsky-2009-unsupervised, VallsVargas2021TowardCR,stammbach-etal-2022-heroes,bamman-etal-2014-bayesian,jahan-finlayson-2019-character}. Others explore authorial decisions in representing characters~\cite{bullard-ovesdotter-alm-2014-computational} or how they evolve over retellings~\cite{besnier-2020-history}.

Some approaches learn character representations directly. \citet{inproceedings} explore word embeddings as character representations. \citet{holgate-erk-2021-politeness} learn vector representations using masked entity prediction as a training objective. Most similar to our work, \citet{inoue-etal-2022-learning} propose a benchmark for evaluating character representations. Their work takes a broad multi-author, multi-task perspective, while ours dives more deeply into characters by a single author, exploring character similarity from three different angles.

\section{A Three-Part Benchmark for Evaluating Character Similarity}

\begin{table*}[t]
\begin{tabular}{l|l}
    Heroines:&Emma Woodhouse, Elizabeth Bennet, Elinor Dashwood, Marianne Dashwood,\\
    &Fanny Price, Catherine Morland, Anne Elliot\\
     Heroes:&George Knightley, Fitzwilliam Darcy, Edward Ferrars, Edmund Bertram, Henry Tilney,\\
     &Frederick Wentworth, Colonel Brandon\\
     Deceivers:& John Thorpe, George Wickham, John Willoughby, William Elliott, Henry Crawford,\\
     &Frank Churchill\\
     Rivals:&Caroline Bingley, Lucy Steele, Louisa Musgrove, Mary Crawford, Harriet Smith\\
     Wooers:&Henry Crawford, William Elliot, Philip Elton, Charles Musgrove, William Collins,\\
     &John Thorpe\\
     Siblings:&Marianne Dashwood, Jane Bennet, Lydia Bennet, Mary Bennet, Kitty Bennet,\\
     &Susan Price, Mary Musgrove, Elizabeth Elliot, Isabella Knightley, James Morland,\\
     &William Price\\
     Parents:&Mr. Bennet, Sir Walter Elliot, Lieutenant Price, Mr. Woodhouse, Mrs. Bennet,\\
     &Mrs. Dashwood, Mrs. Price, Mrs. Morland\\
\end{tabular}
    \caption{Narrative Roles benchmark summary}
    \label{tab:narrative_sum}
\end{table*}

Character similarity is a multi-faceted concept. Two characters may play the same role in a narrative or follow the same plot trajectory. They may have similar personality traits or fill similar social roles. \dataset uses a multi-faceted approach to character similarity that explores three aspects of literary characterhood: shared narrative roles, shared social characteristics, and pairwise comparisons from expert analysis.\footnote{The dataset and support code are available at \ifsub{https://anonymous.4open.science/r/AustenAlike-5A55}\else{https://github.com/Wellesley-EASEL-lab/AustenAlike}\fi.}
The \dataset benchmark focuses on characters from the six Jane Austen novels published within or immediately after her lifetime: \textit{Sense and Sensibility}, \textit{Pride and Prejudice}, \textit{Mansfield Park}, \textit{Emma}, \textit{Persuasion}, and \textit{Northanger Abbey}. We include all named characters who speak more than once, except those who die in the first chapter.%\footnote{Named characters excluded by these criteria include Mr. Dashwood, Sarah (Sally) Morland, and Lady Elliot.}

\subsection{Social Characteristics}

Jane Austen's novels highlight how her character's choices are impacted by their position in society. Although her characters struggle to varying degrees to reconcile their desires with constraints imposed by gender, rank, and wealth, these social characteristics play a large part in determining the options available to them within the novel.

We consider five demographic dimensions that define social relationships within Austen's writing: marital status, gender, rank, age, and wealth. There are other social characteristics that demarcated opportunities within Austen's historical context, such as race and nationality; however, the characters under consideration are homogeneously White and English.\footnote{Given the exclusion of Austen's unfinished \textit{Sanditon}.} A summary of the social categories and the size of each group is in Appendix \ref{app:benchmarks}.

\paragraph{Rank}

Although almost all of Jane Austen's characters belong to the upper middle or lower upper classes, their relative social rank is nonetheless important to their prospects. Most characters are gentry: independently wealthy, often landowners. Lower-ranked characters belong to professions. Following social conventions of the time, an unmarried woman has her father's rank and a married woman her husband's.%\footnote{Female peers retain their titles if they marry someone of inferior rank, but this does not arise in Austen.} 

\paragraph{Wealth}

Austen novels center on questions of wealth, particularly as they relate to marital prospects. As a result, the wealth of unmarried characters is typically stated. The wealth of married characters is not always stated. We draw on estimates from \citet{heldman-economics-1990} and \citet{toran-economics-2015}.  

\paragraph{Gender}

The genders of all Austen characters are overt and stable. All characters are Male or Female.

\paragraph{Age}

Character ages are reasonably stable as almost all plot events take place within a year. If a character's age is not mentioned, we estimate from the ages of their family members. 

\paragraph{Marital status}

Marital status is a key social characteristic of Austen characters. We divide characters into four groups: Single, Married, Widowed, and Transitional, a group comprising the handful of characters whose marital status changes before the climax of the novel. 

\subsection{Narrative Roles}

Another way in which characters can resemble each other is in the role they play in the narrative structure of the work. We define seven narrative roles:

\begin{itemize}
    \item Heroine: each novel has at least one protagonist who is an unmarried woman seeking a marriage partner.
    \item Hero: the character that each protagonist marries at the novel's end.
    \item Deceiver: each novel features a character who sets key events in motion by lying about himself or the heroine. 
    \item Rival: an alternate love interest for the hero.
    \item Wooer: an alternate love interest for the heroine.
    \item Parents: the parents of the heroine.
    \item Siblings: the siblings of the heroine.
\end{itemize}. 

These groupings are shown in Table \ref{tab:narrative_sum}.

\subsection{Wisdom-of-the-Experts Character Pairs}

In our most fine-grained benchmark, we look at characters who have been identified as similar by literary scholars. We use a wisdom-of-the-crowds approach, but with an expert crowd: authors of articles published in \textit{Persuasions}, the Jane Austen Society of North America's peer-reviewed journal.  

We manually reviewed 43 volumes of \textit{Persuasions} to create a set of character pairings. We extract all instances of a similarity or shared property discussed in an article. When an article mentions a similarity between more than two characters, we add all pairings from the set. The resulting dataset contains 5740 character comparison pairs.

The identified comparisons are diverse, encompassing traits from our other benchmarks, such as rank, age, and narrative role, as well as more nuanced commonalities. For instance, \textit{Persuasions} authors describe Edward Ferrars and Frank Churchill as similar because both are secretly engaged; Emma Woodhouse and Lady Catherine de Bourgh because they oversee charitable work; and Isabella Thorpe and Lydia Bennet because of their flirtatiousness. These expert-identified pairings provide a comprehensive view of character similarity.

\section{Building Computational Representations of Character}

We build computational representations of character from the output of two literary pipelines. We construct representations out of the features they extract: for \booknlp, events, quotes, and modifiers; for \fanfic, quotes and assertions.

\subsection{Character Mentions}

We use each pipeline to identify all character mentions, perform coreference resolution, and aggregate character mentions. We then merge and filter character clusters using a handwritten alias map for Austen character names. 

\subsection{Feature Embeddings}

We retrieve contextualized embeddings for each kind of feature. For events and modifiers, which are single words, we retrieve a contextualized embedding of the word in its context using T5 (11B)~\citep{t5}. For quotes and assertions, we retrieve sentence embeddings using NV-Embed (7.85B)~\citep{lee2024nvembed}. We center each kind of feature embedding by subtracting the mean of all embeddings for the feature.

For each feature and character, we construct a character representation by averaging the embeddings of the character's features. For events, we average the character's agent events and patient events separately and concatenate the vectors. This process produces 5 representations per character: an assertion vector, a modifier vector, an event vector, and two quote vectors (one per pipeline). 

\subsection{GPT-4 comparison}

We provide a non-featured-based comparison by querying a pretrained large language model, GPT-4~\citep{gpt4}, for character similarity rankings. Given the popularity of Austen's work, we assume that GPT-4's training data contains all six novels and many web pages discussing them. 

We extract character similarities using three approaches: asking GPT-4 to select the most similar character from a list of all benchmark characters; asking GPT-4 to select the most similar character and explain its choice; and asking GPT-4 to choose the ten most similar characters from a list of all benchmark characters. We repeat each experiment 5 times (further details in Appendix \ref{app:gpt4}).

\section{Evaluating character similarity}

We have proposed three benchmarks that capture different aspects of character similarity. For the social and narrative roles benchmarks, we are interested in the similarity between characters in the same groupings. For the expert benchmark, we are interested in whether characters are most similar to those they are paired with by experts.

\subsection{Grouping evaluation}

The Social and Narrative benchmarks define groupings of characters. We explore how strongly these groupings are captured by computational character representations using two evaluation metrics.

%\paragraph{Cluster Purity} 

%For each character grouping defined in a benchmark, we use K-Means clustering to partition the character representations into two groups, and then assess how well these clusters align with benchmark groupings. We treat each benchmark group as partitioning the character set between group members and non-group members. We compute the average \textit{cluster purity} using the adjusted Rand index clustering comparison metric. A score of zero or below indicates at-chance clustering, while 1 indicates a perfect match between the labels and the clusters.  

%\paragraph{Most Similar Characters}

%We also explore whether the characters that are most similar to each other are within the same character grouping. For each character in a grouping, we compute the cosine similarity of their representation with all other representations. We look at the \textbf{top-k most similar} characters, and report the percentage that belong to the same character grouping.

\paragraph{In-group Cosine Similarity}

We explore whether characters are more similar to characters within their group than those outside of their group. We compute the average cosine similarity between a grouped character and all other group members, and compare it to the average cosine similarity between the character and non-group characters. We call this \textit{in/out-group cosine similarity difference}.

\paragraph{Most Similar Character}

We also ask whether very similar characters come from the same groups. We count how often the single character with highest cosine similarity to the target character belongs to the same group.

\subsection{Pairing evaluation}

For the Expert benchmark, we measure the extent to which the cosine similarities of each kind of representation align with the expert-identified pairs using three metrics:

\paragraph{Correlation} We look at the correlation between cosine similarity of two character representations and the number of times experts describe the two characters as similar. We calculate Pearson's $\rho$ to measure the strength of the correlation. %between the count of expert pairings and the cosine similarity of the paired character representations.

\paragraph{Ranking similarity} 

Literary experts may be more interested in identifying highly similar characters than in quantifying degrees of dissimilarity. We identify the ten most similar characters according experts and to cosine similarity, and compute the alignment between the lists using Jaccard similarity. Jaccard similarity measures the intersection of the groups divided by their union. If the two lists are completely different, their Jaccard similarity is 0; if they mostly agree, it is close to 1. %This is a more appropriate metric than alternatives for ranked data, like the Kendall correlation coefficient, because the rankings we compare may contain different characters. 

\paragraph{Top character in ten-most similar}

Finally, we focus on the top expert-identified pairings. We count how often the character who experts pair most with a target character has one of the ten highest cosine similarities to the target character.

%\subsection{Across Novel Comparisons}

%Because the character representations we explore are extracted from textual features, the computational representations of characters within the same novel are likely to be most similar. For instance, characters in \textit{Northanger Abbey} are more likely to engage in reading events since this is a theme of the novel. We therefore compute two versions of each metric: one that includes all characters, and one that includes only characters in other novels. 

\begin{figure*}[t]
    \centering
    \includegraphics[width=\textwidth]{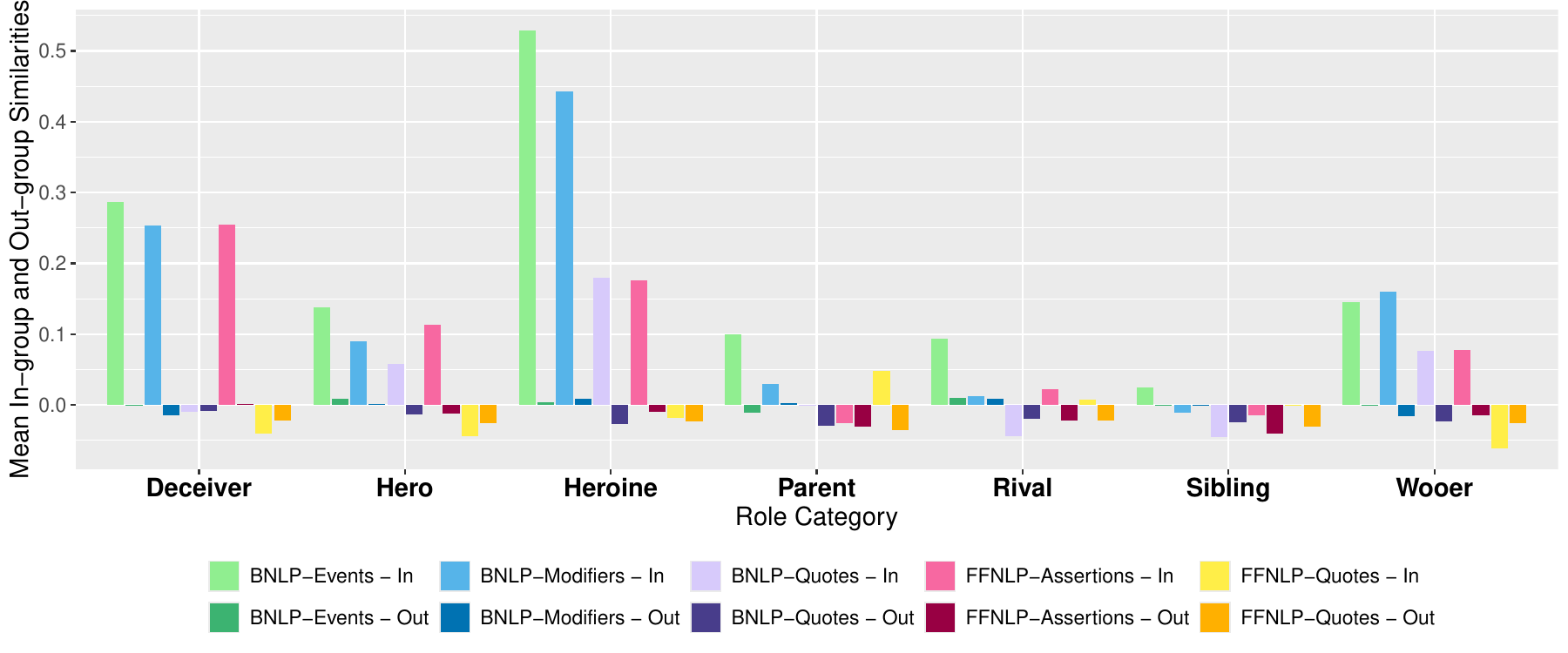}
    \caption{Narrative Role Benchmark: Mean cosine similarities between same-group characters and other characters by representation type.}
    \label{fig:simdiffs}
\end{figure*}

\begin{table*}[t]
    \centering
    \begin{tabular}{|l|l|l|l|l|l|l|l|}\hline
        System & Hero&Heroine&Deceiver&Rival&Wooer&Parent&Sibling \\\hline
        \fanfic Assertions&0.29&0.43&0.33&0&0&0.18&\textbf{0.29}\\
        \booknlp Events&0&\textbf{1}&0.36&0.09&0.18&\textbf{0.35}&0 \\
        \booknlp Modifiers&0&0.86&0.33&0.2&0&0.27&0.18 \\
        \booknlp Quotes&0.13&0.78&0.57&\textbf{0.33}&0.43&0.08&0 \\
        \fanfic Quotes&0&0.43&0&0.14&0&0.18&0.08\\\hline
        GPT-4&0.43&0.43&0.5&0&0&0.33&0.25\\
        GPT-4 Reasoning&\textbf{0.86}&\textbf{1}&\textbf{0.83}&0.17&\textbf{0.5}&0.42&0.08\\\hline
    \end{tabular}
    \caption{Narrative Role Benchmark: Average occurrence of most similar character in same narrative role group by character representation. Characters from same novel are excluded.}
    \label{tab:narr_topsim}
\end{table*}

\section{Results}

We explore how well computational representations of character capture aspects of character similarity using the three-part \dataset benchmark.

\subsection{Narrative Roles Benchmark}

The narrative roles benchmark explores similarity between characters who play similar roles in the plot of a novel. Are heroines similar to other heroines? Are parents similar to other parents? If parents are described similarly to other parents, assertion- and modifier-based representations should capture their similarity; if they say and do similar things as other parents, their quote- and event-based representations should be similar.

\subsubsection{Are same-role characters more similar?}

We test whether characters who share the same narrative role are more similar than characters who do not. We compare the average cosine similarity of representations within a narrative role group to their similarity to non-group members. We compute the in-group and out-group scores for each character in a target role group and average them. 

Figure \ref{fig:simdiffs} plots the cosine similarity for characters within the same narrative role group compared to characters outside of the group. We observe that event- and assertion-based representations are the best at showing dissimilarity for characters outside of the role group. The \fanfic quote-based representations show the weakest differences between in-group and out-group members.

\subsubsection{Is the most similar character from the same group?}

We also explore whether a target character's most similar character belongs to the same narrative role group. For each character, we count how often the character with highest cosine similarity belongs to the same role group. Feature-based representations can be skewed towards same-novel similarity: for instance, characters in \textit{Northanger Abbey} are more likely to engage in reading events since this is a theme of the novel. We therefore explore results with and without characters  from the same novel.

Table \ref{tab:narr_topsim} reports how often the most similar character occurs in the same role group, with same-novel characters excluded (inclusive version in Appendix \ref{app:results}). We see marked differences between categories. Heroines are frequently similar to heroines for all representations, while other groups have lower rates of same-group membership.  

The \booknlp quote representations capture narrative role similarity better than the \fanfic quote representations, perhaps because \booknlp is trained on literary fiction. However, \fanfic assertions perform competitively in two of the most challenging categories for feature-based representations, Hero and Sibling.

We observe that GPT-4, when asked to justify its decision, is more sensitive to narrative role than the feature-based representations in about half of the categories. However, without reasoning-prompting, it is no better than the feature-based representations, identifying selecting a heroine as the most similar to heroines only 43\% of the time.

Qualitatively, a challenging aspect of this benchmark seems to stem from young single characters with different narrative roles. Like heroes and heroines, deceivers, wooers, and rivals tend to be unmarried and of a similar age. We observe that heroes tend to be similar to deceivers (10/69 out-group cases) and vice versa (12/50 out-group cases), and rivals to heroines (26/64) and vice versa (6/31 out-group cases), aligning with the social characteristics of each set. The error patterns for the remaining categories seem less clear, perhaps reflecting the limited mentions of parent characters and the more heterogeneous characteristics of siblings.

\begin{figure*}[t]
    \centering
    \includegraphics[width=\textwidth]{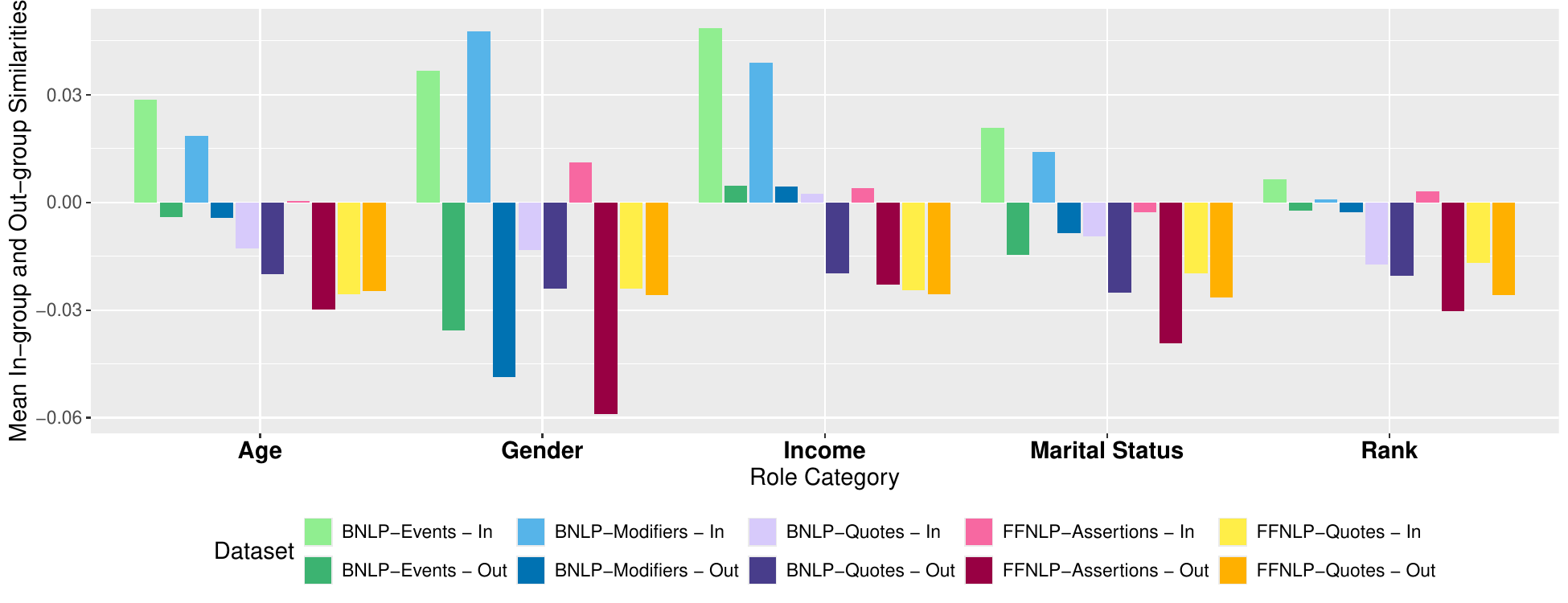}
    \caption{Social Benchmark: average differences in cosine similarity between same-group characters and other characters by character representation and social role group.}
    \label{fig:soc_simdiffs}
\end{figure*}

\begin{table*}[t]
    \centering
    \begin{tabular}{|l|l|l|l|l|l|}\hline
        System & Age&Gender&Income&Marital Status&Rank\\\hline
        \fanfic Assertions&0.16&0.9&0.02&0.5&0.34\\
        \booknlp Events&0.23&0.76&0.07&0.51&0.29\\
        \booknlp Modifiers&0.22&0.80&0.05&0.46&0.19\\
        \booknlp Quotes&0.06&0.63&0.15&0.42&0.26\\
        \fanfic Quotes&0.13&0.54&0.02&0.3&0.25\\\hline
        GPT-4&0.26&0.80&\textbf{0.21}&0.52&\textbf{0.42}\\
        GPT-4 Reasoning&\textbf{0.32}&\textbf{0.98}&0.07&\textbf{0.58}&0.39\\\hline
    \end{tabular}
    \caption{Social Benchmark: average occurrence of most similar characters in the same social group by character representation. Characters from same novel are excluded.}
    \label{tab:soc_topsim}
\end{table*}

\subsection{Social Benchmark}

The second \dataset benchmark evaluates character similarity on the basis of social characteristics. It groups characters based on five demographic features: rank, wealth, gender, age, and marital status. Modifiers and assertions may directly describe these characters. However, given that a character's social status delimits the set of actions and utterances available to them, we also expect event- and quote-based representations to echo back similarities based on these characteristics.

\subsubsection{How similar are characters with shared social characteristics?}

We explore whether characters within the same group in each of the social categories are most similar to each other. Figure \ref{fig:soc_simdiffs} plots the average cosine similarity for characters within the same social group compared with non-group members.

We observe that the event-based representations are the most reliable for distinguishing social similarity. Gender shows the sharpest in-group/out-group differences for all three categories, followed by income. Quote-based representations struggle to capture similarity by social group: the \fanfic quote-based representations do not capture differences for any of the criteria, while the \booknlp quote-based representations show only a (weak) in-group/out-group difference for income.

\subsubsection{Is the most similar character from the same group?}

We also focus more narrowly on the top-most similar character. Table \ref{tab:soc_topsim} shows how often the character with the highest cosine similarity to the target character occurs in the same social group. Top character representations most commonly share gender and then marital status. This makes sense, since Austen's plots center around courtship: these key aspects of identity should be reflected in how they are described and the events they participate in.

GPT-4's similarity judgments align with social characteristics more strongly than any of the feature-based representations. Quote-based representations do not seem to capture similarity by social characteristics as well as the other feature-based representations in most categories. 

\begin{table*}[t]
    \centering
    \begin{tabular}{|l|l|l|l|}\hline
    Dataset&Pearson's $\rho$&Jaccard Similarity&Top in Top 10\\\hline
\fanfic Assertions&0.29&\textbf{0.03}&\textbf{0.69}\\
\booknlp Events&\textbf{0.4}&0.02&0.34\\
\booknlp Modifiers&0.28&0.01&0.29\\
\booknlp Quotes&0.27&\textbf{0.03}&0.56\\
\fanfic Quotes&0.15&0.02&0.49\\\hline
GPT-4&-&-&0.52\\
GPT-4 Reasoning&-&-&0.56\\
GPT-4 Top Ten List&-&0.02&-\\\hline
    \end{tabular}
    \caption{Expert Benchmark: measures of alignment between expert pairing counts and computational similarity.}
    \label{tab:expert}
\end{table*}

\subsection{Expert Benchmark}

Our last benchmark takes an expert wisdom-of-the-crowd approach. The expert benchmark contains counts of character similarity pairings. We compare these pairing counts to the cosine similarity between the computational representations of the two characters to evaluate how well computational representations aligns with expert judgments of character similarity.

\subsubsection{Does cosine similarity correlate with expert judgments?} 

We examine how well computational character representations align with expert judgments by measuring the correlation between expert character pairings and cosine similarity. We posit that high quality computational representations should produce higher cosine similarity between the characters that are more frequently deemed similar by experts.

Table \ref{tab:expert} shows the correlation between expert pairing counts and cosine similarity for each of the computational representations. 

Overall, we observe moderate positive correlations between the cosine similarity of character representations and the number of expert similarity pairings. The \booknlp event representations correlate most strongly with expert pairings, while the \fanfic quote-based representations correlate less strongly than other feature-based representations. This converges with our social and narratological similarity findings. 

Although the expert benchmark is useful in differentiating among feature-based representations, it is also important to note that none of the feature-based representations are strongly correlated with expert judgments. This shows that there are many aspects of character similarity that are apparent to human readers that remain uncaptured in the computational character representations we explore.

\subsubsection{Is there agreement on the most similar characters?}

Correlations between cosine similarity and expert pairing counts may be skewed by very dissimilar characters, whose expert pairings are few. We also look at two measures of agreement for the most similar characters.

For each character, we retrieve the ten characters with the highest cosine similarity, and the ten characters with whom they are most frequently paired by experts. We then measure agreement by computing the Jaccard similarity of the two sets. 

Table \ref{tab:expert} shows the average Jaccard similarity these top ten sets. The Jaccard scores are uniformly low, indicating that cosine similarity tends not to identify the same set of highly similar characters as experts. Interestingly, GPT-4 does not appear any more successful at identifying expert-aligned similar characters than the feature-based approaches, despite its success in identifying socially and narratologically similar characters.

We also examine how often the single character that experts compare most to a target character occurs within the target's top ten closest representations by cosine similarity. Table \ref{tab:expert} shows the average success on this lenient measure. 

Even with this easier measure, the expert benchmark is quite challenging. GPT-4 includes the expert top character in its top ten list only half of the time. The best feature-based representation, \fanfic assertions, include it 69\% of the time. Since this is a very lenient measure of success, this illustrates the large gaps that remain between similarity by computational representations of character, pretrained LLM understanding of character similarity, and expert evaluations.

\section{Conclusion}

We present \dataset, a three-part Jane Austen benchmark for evaluating multiple aspects of character similarity: narrative role similarity, social similarity, and expert judgments of character similarity drawn from prior scholarly analysis. We use \dataset to evaluate five computational representations of character built atop features extracted by pipelines for analyzing English literature. 

We find that event- and assertion-based representations tend to capture character similarity better than quote-based representations. Overall, however, our results show how much work still remains to be done to improve computational representations of character: feature-based representations and GPT-4 alike struggle to place the expert-identified most similar character in their top ten lists of character similarity. We hope that by providing a multi-faceted benchmark with expert judgments, \dataset can guide future work on computational representations of character.

\section*{Limitations}

We have evaluated five kinds of feature-based character representations across two systems. However, our approach has a number of limitations.

\paragraph{Noisy Character Data}
Both pipelines produce character clusters with some amount of inconsistency and error. In some cases, the pipelines failed to resolve multiple ways of referring to the same character (\textit{Miss Tilney}, \textit{Eleanor Tilney}). We post-process the output with an Austen-specific alias map; to extend our work to other works of literature, this post-processing step would need to be manually extended.

\paragraph{Missing characters} 

Both pipelines failed to extract features for some characters included in our benchmark. \booknlp failed to identify twelve characters and \fanfic failed to identify four. This was most impactful in the siblings and parents subsets of the narrative roles benchmark.

\paragraph{Generalizability}  Our benchmark focuses on characters from the work of Jane Austen. As a result, it may favor methods of deriving computational representations that are trained on similar literary text. This may affect our comparison of \fanfic and \booknlp quotes, as noted above.

\section*{Ethics Statement}

Our work does not involve any human data. The literary works we analyze are in the public domain. The computational resources involved in our experiments are also modest: all contextualized embeddings were extracted using less than 12 hours on a single Nvidia RTX A6000 GPU.

%\section*{Acknowledgements}

%We would like to thank Eni Mustafaraj, Peter Mawhorter, and Yoon Sun Lee for their helpful comments on this work.

\bibliography{anthology,custom}
\bibliographystyle{acl_natbib}

\appendix

\section{Further Details of Benchmark Construction}\label{app:benchmarks}

\subsection{Social Benchmark}

\paragraph{Rank} To achieve a more even balance across groups, we partition untitled gentry into two groups: New Gentle, characters whose fathers were not gentlemen, and Gentle, representing more established gentry. We consolidate professional characters into three groups: a military group encompassing the army and navy; a professional group encompassing business, law, and farming; and a clergy group. This totals six categories: New Gentle, Gentle, Gentry, Military, Profession, Clergy, and Nobility.

\begin{table}[t]
    \centering
    \begin{tabular}{|l|l|l|}\hline
    Category&Group&N\\\hline
    Rank&Nobility&2\\
        &Titled Gentry&15\\
        &Gentle&48\\
        &New Gentle&5\\
        &Clergy&12\\
        &Military&13\\
        &Profession&14\\\hline
    Wealth&\pounds 50&8\\
        &\pounds 51-\pounds 250&7\\
        &\pounds 251-\pounds 500&9\\
        &\pounds 501-\pounds 1000&8\\
        &\pounds 1001-\pounds 3000&6\\
        &\pounds 3001+&5\\\hline
    Gender&Male&50\\
         &Female&59\\\hline
    Age&< 18&8\\
        &18-20&13\\
        &21-24&16\\
        &25-27&18\\
        &28-30&12\\
        &31-40&13\\
        &41-50&19\\
        &51+&10\\\hline
    Marital Status&Single&48\\
        &Transitional&6\\
        &Married&42\\
        &Widowed&13\\\hline
    \end{tabular}
    \caption{Social Characteristics benchmark summary}
    \label{tab:social_sum}
\end{table}

\paragraph{Wealth} Wealth for women is generally reported as a total sum, while men's fortunes are typically stated in terms of yearly income. We convert all figures to yearly incomes assuming the 5\% yearly dividend standard during Austen's time~\citep{toran-economics-2015}. %One challenge in grouping characters by wealth is the fact that some characters' fortunes change over the course of the novel. We extract two kinds of wealth features, when available, for each character: starting income and ending income. 

\paragraph{Marital Status} Marital status tends to remain stable until the end of each novel: although many single characters marry, most marriages take place in the last chapter. 

\subsection{Narrative Roles Benchmark}

\paragraph{Heroines}

All Jane Austen novels involve young people finding marriage partners. Each novel has at least one protagonist who is an unmarried woman seeking a marriage partner. \textit{Sense and Sensibility} focuses on a pair of sisters who both marry by the end of the novel; we treat both as protagonists/heroines. Heroines should be particularly easy to distinguish from other narrative roles since they are the main viewpoint characters in Austen's novels. 

\paragraph{Heroes}

We use the term \textit{hero} for the character that each protagonist marries at the novel's end.

\paragraph{Deceiver}

Each of Austen's novels features at least one character who lies in a way that sets key events in motion. Frequently, this character misrepresents himself to the heroine in a key way (Wickham in \textit{Pride and Prejudice}; Willoughby in \textit{Sense and Sensibility}); in other cases, the character lies to conceal an ulterior motive (William Elliot in \textit{Persuasion}; Frank Churchill in \textit{Emma}). In one case, this character spreads lies about the heroine herself (John Thorpe in \textit{Northanger Abbey}).

\paragraph{Rivals and Wooers}

In each of the six novels, there is at least one character who serves as a rival, an alternate love interest for the hero. In all but one novel (\textit{Sense \& Sensibility}), there is a character who unsuccessfully courts the heroine; we refer to these characters as \textit{wooers}.

\paragraph{Family roles}

Austen's novels are concerned with domestic settings and interactions within a relatively confined society. As a result, there are numerous family members. We look at two groups: parents and siblings. In the case of \textit{Mansfield Park}, in which the heroine is raised in her uncle's family, we considered including her guardians but excluded them to be consistent with other mentors (Lady Russell in \textit{Persuasion}) and temporary guardians (the Allens in \textit{Northanger Abbey}). 

\section{Further Details of GPT-4 Experiments}\label{app:gpt4}

We run three experiments to extract character similarities from GPT-4: a top character experiment, a top character experiment with reasoning, and a top ten characters experiment. We run each experiment five times at temperature=0.2.

The prompts are shown below (full list of characters omitted for readability). \textit{c} represents the name of the target character, and \textit{cIndex} is that character's number in the list.\\

\noindent \textbf{Top Character Prompt}

\noindent Consider the following list of Jane Austen characters:

\noindent 1. Anna Weston\\
2. Augusta Elton\\
...\\
108. Sir John Middleton\\
109. Thomas Palmer\\

\noindent Which character is \textit{c} most similar to (other than \textit{c})? Respond with only a number. Do not choose \textit{cIndex}.\\

\noindent \textbf{Top Character with Reasoning Prompt}

\noindent Consider the following list of Jane Austen characters:

\noindent 1. Anna Weston\\
2. Augusta Elton\\
...\\
108. Sir John Middleton\\
109. Thomas Palmer\\

\noindent Which character is \textit{c} most similar to (other than \textit{c})? Describe your reasoning and then reply with the number of the character. Do not choose \textit{cIndex}.\\

\noindent \textbf{Top Ten Characters Prompt}

\noindent Consider the following list of Jane Austen characters:

\noindent 1. Anna Weston\\
2. Augusta Elton\\
...\\
108. Sir John Middleton\\
109. Thomas Palmer\\

\noindent List the 10 characters that are most similar to \textit{c} (other than \textit{c}). Consider characters from all Austen novels. Reply with just their numbers. Do not choose \textit{cIndex}.

\section{Further Results}\label{app:results}

\subsection{Narrative Role Benchmark}

Table \ref{tab:narr_topsim2} shows how often the most similar character is within the same narrative role set as the target character, with all books included. Table \ref{tab:narr_topsim} excludes characters from the same book.

\begin{table*}[t]
    \centering
    \begin{tabular}{|l|l|l|l|l|l|l|l|}\hline
        System & Hero&Heroine&Deceiver&Rival&Wooer&Parent&Sibling \\\hline
        \fanfic Assertions&0.14&0.36&0.17&0&0&0.18&0.25\\
        \booknlp Events&0.07&1&0.33&0.08&0.17&0.36&0 \\
        \booknlp Modifiers&0&0.86&0.33&0.25&0&0.27&0.18 \\
        \booknlp Quotes&0.07&0.64&0.33&0.17&0.25&0.09&0 \\
        \fanfic Quotes&0.14&0.21&0&0.08&0&0.14&0.08\\\hline
        GPT-4&0.43&0.43&0.5&0&0&0.33&0.25\\
        GPT-4 Reasoning&0.86&1&0.83&0.17&0.5&0.42&0.08\\\hline
    \end{tabular}
    \caption{Narrative Role Benchmark: Average occurrence of most similar character in same narrative role group by character representation. Characters from same novel are included.}
    \label{tab:narr_topsim2}
\end{table*}

\subsection{Social Benchmark}

Table \ref{tab:soc_topsim2} shows how often the most similar character is within the same social role set as the target character, with all books included. Table \ref{tab:soc_topsim} excludes characters from the same book.

\begin{table*}[t]
    \centering
    \begin{tabular}{|l|l|l|l|l|l|}\hline
        System & Age&Gender&Income&Marital Status&Rank\\\hline
        \fanfic Assertions&0.18&0.75&0.13&0.52&0.41\\
        \booknlp Events&0.23&0.77&0.13&0.51&0.30\\
        \booknlp Modifiers&0.21&0.78&0.07&0.46&0.19\\
        \booknlp Quotes&0.09&0.58&0.15&0.40&0.34\\
        \fanfic Quotes&0.10&0.49&0.05&0.37&0.34\\\hline
        GPT-4&0.26&0.80&\textbf{0.21}&0.52&\textbf{0.42}\\
        GPT-4 Reasoning&\textbf{0.32}&\textbf{0.98}&0.07&\textbf{0.58}&0.39\\\hline
    \end{tabular}
    \caption{Social Benchmark: average occurrence of most similar characters in the same social group by character representation. Characters from same novel are included.}
    \label{tab:soc_topsim2}
\end{table*}

\subsection{Expert Benchmark}

Tables \ref{tab:pair_cor} and \ref{tab:pair_cor2} shows Pearson's $\rho$ correlations between cosine similarity and expert pairing counts by novel, with characters from the same novel included and excluded respectively.

\begin{table*}[t]
    \centering
    \begin{tabular}{|l|l|l|l|l|l|l|l|}\hline
         Novel & \textit{Emma}&\textit{MP}&\textit{NA}&\textit{Pers.}&\textit{P\&P}&\textit{S\&S}&All\\\hline
         \fanfic Assertions&0.30&0.38&0.29&0.27&0.23&0.28&0.29\\
\booknlp Events&0.44&0.44&0.43&0.31&0.37&0.43&\textbf{0.4}\\
\booknlp Modifiers&0.31&0.31&0.25&0.26&0.26&0.29&0.28\\
\booknlp Quotes&0.26&0.28&0.28&0.20&0.24&0.35&0.27\\
\fanfic Quotes&0.21&0.20&0.15&0.11&0.10&0.11&0.15\\\hline
    \end{tabular}
    \caption{Expert Benchmark: Pearson's $\rho$ correlation between cosine similarity and expert pairing count by character representation. Character pairs with no expert mentions are excluded.}
    \label{tab:pair_cor}
\end{table*}

\begin{table*}[t]
    \centering
    \begin{tabular}{|l|l|l|l|l|l|l|l|}\hline
         Novel & \textit{Emma}&\textit{MP}&\textit{NA}&\textit{Pers.}&\textit{P\&P}&\textit{S\&S}&All\\\hline
         \fanfic Assertions&0.3&0.38&0.34&0.23&0.33&0.27\\
\booknlp Events&0.47&0.48&0.50&0.47&0.45&0.48\\
\booknlp Modifiers&0.34&0.37&0.33&0.38&0.34&0.35\\
\booknlp Quotes&0.16&0.11&0.27&0.13&0.28&0.27\\
\fanfic Quotes&-0.01&0.04&0.15&0.04&0.04&-0.07\\\hline
    \end{tabular}
    \caption{Expert Benchmark: Pearson's $\rho$ correlation between cosine similarity and expert pairing count by character representation. Characters from the same novel are excluded. Character pairs with no expert mentions are excluded.}
    \label{tab:pair_cor2}
\end{table*}

\end{document}